\documentclass[runningheads]{llncs}
\usepackage[T1]{fontenc}
\usepackage{graphicx}
\usepackage{multicol}
\usepackage{multirow}
\usepackage{amsmath}
\usepackage{booktabs}
\usepackage[misc]{ifsym}
\usepackage{bbold}
\begin{document}
\title{Semantic Consistency Regularization with Large Language Models for Semi-supervised \\ Sentiment Analysis}
\titlerunning{Semantic Consistency Regularization with Large Language Models}
%
\author{Kunrong Li\inst{1} \and
Xinyu Liu\inst{2} \and
Zhen Chen$^{\text{\Letter}}$\inst{3}}
\authorrunning{Li, Liu, and Chen}
%
%
\institute{School of Physical and Mathematical Sciences, Nanyang Technological University, Singapore \and
Department of Electronic Engineering, The Chinese University of Hong Kong, Hong Kong SAR \and
Centre for Artificial Intelligence and Robotics (CAIR), Hong Kong Institute of Science and Innovation, Chinese Academy of Sciences, Beijing, China\\\email{zhen.chen@cair-cas.org.hk}}
\maketitle
\begin{abstract}
Accurate sentiment analysis of texts is crucial for a variety of applications, such as understanding customer feedback, monitoring market trends, and detecting public sentiment. However, manually annotating large sentiment corpora for supervised learning is labor-intensive and time-consuming. Therefore, it is essential and effective to develop a semi-supervised method for the sentiment analysis task. Although some methods have been proposed for semi-supervised text classification, they rely on the intrinsic information within the unlabeled data and the learning capability of the NLP model, which lack generalization ability to the sentiment analysis scenario and may prone to overfit. Inspired by the ability of pretrained Large Language Models (LLMs) in following instructions and generating coherent text, we propose a Semantic Consistency Regularization with Large Language Models (SCR) framework for semi-supervised sentiment analysis. We introduce two prompting strategies to semantically enhance unlabeled text using LLMs. The first is Entity-based Enhancement (SCR-EE), which involves extracting entities and numerical information, and querying the LLM to reconstruct the textual information. The second is Concept-based Enhancement (SCR-CE), which directly queries the LLM with the original sentence for semantic reconstruction. Subsequently, the LLM-augmented data is utilized for a consistency loss with confidence thresholding, which preserves high-quality agreement samples to provide additional supervision signals during training. Furthermore, to fully utilize the uncertain unlabeled data samples, we propose a class re-assembling strategy inspired by the class space shrinking theorem. Experiments show our method achieves remarkable performance over prior semi-supervised methods. For example, with 200 labeled data per class, SCR-EE achieves a remarkable performance of 76.13\% accuracy, outperforming previous method FixMatch by 3.42\%.

\keywords{Sentiment analysis, Semi-supervised learning, Large language models, Consistency regularization}
\end{abstract}
\section{Introduction}
One major obstacle in training accurate sentiment analysis models is the lack of large-scale labeled datasets. Manually annotating sentiment corpora is a labor-intensive and time-consuming process, which often requires the domain expertise to capture the subtle distinction of sentiment expression in contexts. This data scarcity has hindered the development of supervised learning models, which typically need abundant labeled data for effective training. To overcome this limitation, semi-supervised learning approaches have emerged as a promising solution. By leveraging a large number of unlabeled data and limited labeled data, semi-supervised methods can potentially learn more generalizable representations and mitigate the reliance on extensive manual annotations of the sentiment data. Semi-supervised learning \cite{duarte2023review} have shown notable data efficiency in text classification tasks. They often employ a consistency regularization framework that minimizes the predictions of unlabeled data of different augmentations \cite{liu2024diffrect}. However, the traditional data augmentation techniques in NLP, such as back-translation or simple word replacements, often rely on shallow heuristics or statistical methods that lack a deep understanding of language semantics. While these techniques can introduce some variations in the data, they are insufficent in two aspects: (1) They alter the text at a surface level, potentially introducing noise or inconsistencies that could hinder the model's ability to learn robust representations. (2) These methods rely solely on the unlabeled data and the learning capability of the model, which may lack the generalization ability required for sentiment analysis and can be prone to overfitting or even performance collapse. 
 
Recently, Large Language Models (LLMs) like GPT-3, LLaMA \cite{touvron2023llama4}, PaLM \cite{anil2023palm5}, and Qwen \cite{qwen} have demonstrated very impressive language understanding and generation capabilities, revolutionizing various natural language processing tasks. These LLMs require massive amounts of textual data for training, thus have acquired a deep understanding of language patterns, semantics, and world knowledge. They can not only generate fluent and contextually relevant text but also follow specific instructions and prompts to perform controlled text generation tasks. For example, when prompted with appropriate instructions, the pretrained LLMs can rephrase sentences while preserving the core meaning, adjust the sentiment or tone of a given text, or incorporate specific entities or concepts into the generated output \cite{liu2023trustworthy7}.
 
Inspired by the capabilities of the LLMs, we are willing to explore their potential in semi-supervised sentiment analysis. Therefore, we propose a \textbf{S}emantic \textbf{C}onsistency \textbf{R}egularization with Large Language Models (SCR) framework that utilizes LLMs to augment unlabeled text data with semantically consistent variations. Specifically, we introduce two approaches for semantic enhancement using LLMs. The first approach is Entity-based Enhancement (SCR-EE),  which first prompt the LLM to aware of the keywords and components in the original sentence by extracting entities and numerical information from it, then let the LLM to reconstruct the text based on the extracted key information, which could preserve the core semantics implicitly. The second approach is Concept-based Enhancement (SCR-CE), which directly queries the LLM with the original sentence and instruct the model to generate sentences with semantically consistent variations. Afterwards, we utilize a consistency regularization framework, where a consistency loss with confidence thresholding is employed to selectively incorporate high-quality agreement samples as additional supervision signals during model training. Finally, to further leverage the less-confident data sample, inspired by the class-space shrinking theorem \cite{yang2023shrinking}, we propose a class re-assemble method by removing the confusing sentiment class. By enforcing the consistency between the predictions on the original and semantically enhanced samples, our SCR framework encourages the model to learn robust and generalizable representations. The proposed SCR leverages the power of LLMs of capturing the nuanced sentiment expressions in the social media and customer comment texts, to semantically enhance the unlabeled data. Therefore, more informative and diverse data are generated for effective semi-supervised learning. Our contributions are summarized as below:
\begin{itemize}
    \item We propose a novel Semantic Consistency Regularization with Large Language Models (SCR) framework for semi-supervised sentiment analysis. It leverages the remarkable capabilities of pretrained LLMs to semantically enhance unlabeled data, and use consistency loss for efficient model training.
    \item We introduce two approaches within the SCR framework for semantic data augmentation using LLMs: Entity-based Enhancement (SCR-EE) and Concept-based Enhancement (SCR-CE). SCR-EE extracts entities and numerical information from the text and queries the LLM to reconstruct the textual information while preserving semantics, while SCR-CE directly queries the LLM with the original sentence to generate semantic consistent variations.
    \item We propose a class re-assemble method for sentiment analysis to better leverage the less confident data samples, which further enhances the learning efficiency for unlabeled data.
    \item We conduct comprehensive experiments on two sentiment analysis datasets, our SCR achieves the state-of-the-art performance on varied labeling regimes, and surpasses previous methods remarkably. 
\end{itemize}

\section{Related Work}

\subsection{Semi-supervised Learning}
SSL aims to leverage a limited set of annotated samples along with a large collection of unlabeled data. The most representative method is Pseudo-labeling \cite{lee2013pseudo8} which uses a model trained on the labeled data to predict pseudo-labels for the unlabeled data, and use the pseudo-labels for supervision. However, utilizing hard pseudo-labels individually can potentially cause the confirmation biases of the model \cite{arazo2020pseudo9,liu2023decoupled10}, which may affect the model capability or even performance collapse. To mitigate this issue, researchers have demonstrated the advantages of employing consistency regularization approaches \cite{sohn2020fixmatch,abuduweili2021adaptive12,saito2021openmatch13,tarvainen2017mean14}. Specifically, these methods enforce the model to make consistent predictions under different perturbations, which can be applied on input level or model level. For example, Mean-teacher \cite{tarvainen2017mean14} maintains an exponentially moving average of the model weights as the teacher model, and enforces consistency between the student and teacher model outputs on the unlabeled data under input noise perturbations. Recently, researchers have been exploring the perturbation design on input data. One prominent example is FixMatch \cite{sohn2020fixmatch}, which introduces two forms of perturbations which are weak and strong augmentations. The key idea is to leverage the current model's confident predictions on the weakly augmented inputs as pseudo-labels, while enforcing consistency between the strongly augmented version and the pseudo-label. ShrinkMatch \cite{yang2023shrinking} proposed a shrinking strategy for the original class space by adaptively detecting and removing confusion ones for the top-1 class. FreeMatch \cite{wang2022freematch15} adjusted the confidence threshold according to the per-class learning status.

\subsection{Sentiment Analysis}
Sentiment analysis identifies and extracts subjective information in
social media texts and company reports. Accurate sentiment analysis can provide valuable insights for market prediction, financial decision-making, and customer monitoring. Early works in sentiment analysis focused on lexicon-based techniques \cite{sohangir2018financial16}, which use predefined lexicons and compute the overall sentiment score of each sentence according to the occurrence of the words in the lexicons. Despite their straightforward manner, they may fail to adopt to complex words and continuously updating Internet languages. More recent studies have explored the use of deep learning techniques, i.e., utilizing advanced pretrained models such as Transformer like BERT \cite{devlin2018bert17} or GPT-2. For example, Prottasha et al.\cite{prottasha2022transfer18} compared the performance of Word2Vec, GloVe, FastText, and BERT, showing that the BERT models are the optimal model for text representation and could be effective in sentiment analysis. Mutinda et al. \cite{mutinda2023sentiment19} presented a sentiment classification network which combines sentiment lexicon, BERT, and convolutional neural networks. Despite their effectiveness, their fully-supervised training paradigm still requires a large amount of labeled data, which can be expensive and time-consuming to obtain. In this end, some researchers have been developing semi-supervised text classification methods. Zou and Caragea \cite{zou2023jointmatch} proposed a JointMatch method that used two networks with different augmentations to provide pseudo labels for each other, and weighed more disagreement data to keep the two models diverged for effective mutual learning. However, existing text augmentations can introduce noise and hinder robust representation learning, while solely relying on the BERT model to understand the inherent semantics in the unlabeled data may lack generalization and risk to overfitting.

\subsection{Large Language Models for Sentiment Analysis}
With the advent of large language models (LLMs) such as GPT-3, Qwen, LLaMA \cite{touvron2023llama4}, PaLM \cite{anil2023palm5}, and Mistral \cite{jiang2023mistral6}, they have shown impressive performance on a broad range of language tasks in a zero-shot manner. Moreover, as LLMs with the transformer architecture often have large parameters \cite{liu2023efficientvit}, their training-free zero-shot paradigm has a great potential for being employed for sentiment analysis problems. However, existing works have demonstrated that if we direct use them to predict sentiment, they fail to show more superior performance with BERT, and may generate some contradictory or unreasonable responses \cite{zhong2023can21}. LLMs tend to be inferior in complicated tasks that require deeper understanding of the structured sentiment data \cite{zhang2023sentiment}. They are also less accurate in analyzing sentiment information \cite{wang2023chatgpt22}. Despite their inefficiencies in directly giving sentiment results, they have strong abilities in following human instructions and generating additional corpus for data augmentation \cite{hu2024llm23}. Inspired by this, we can leverage LLMs to generate augmented sentence information, and combine with existing SSL consistency frameworks to encourage the model to learn more robust and diverse representations.

\section{Methodology}
We introduce the semi-supervised sentiment analysis task and notations. Specifically, the training dataset consists of two parts: a set of labeled data with $m_l$ samples  $D_l=\{S_i^l,y_i^l\}_{i=1}^{m_l}$, and an unlabeled data set with $m_u$ samples  $D_u=\{S_i^u\}_{i=1}^{m_u}$.  $S$ represents the sentence while the $y$ denotes the corresponding categorical sentiment label. Typically, only a small fragment of data is labeled i.e., $m_u \gg m_l$. The goal is to perform supervised training with $D_l$ and unsupervised training with $D_u$ to achieve good sentiment analysis performance on the test dataset.

\begin{figure*}[t]
    \centering
    \includegraphics[width=0.97\textwidth]{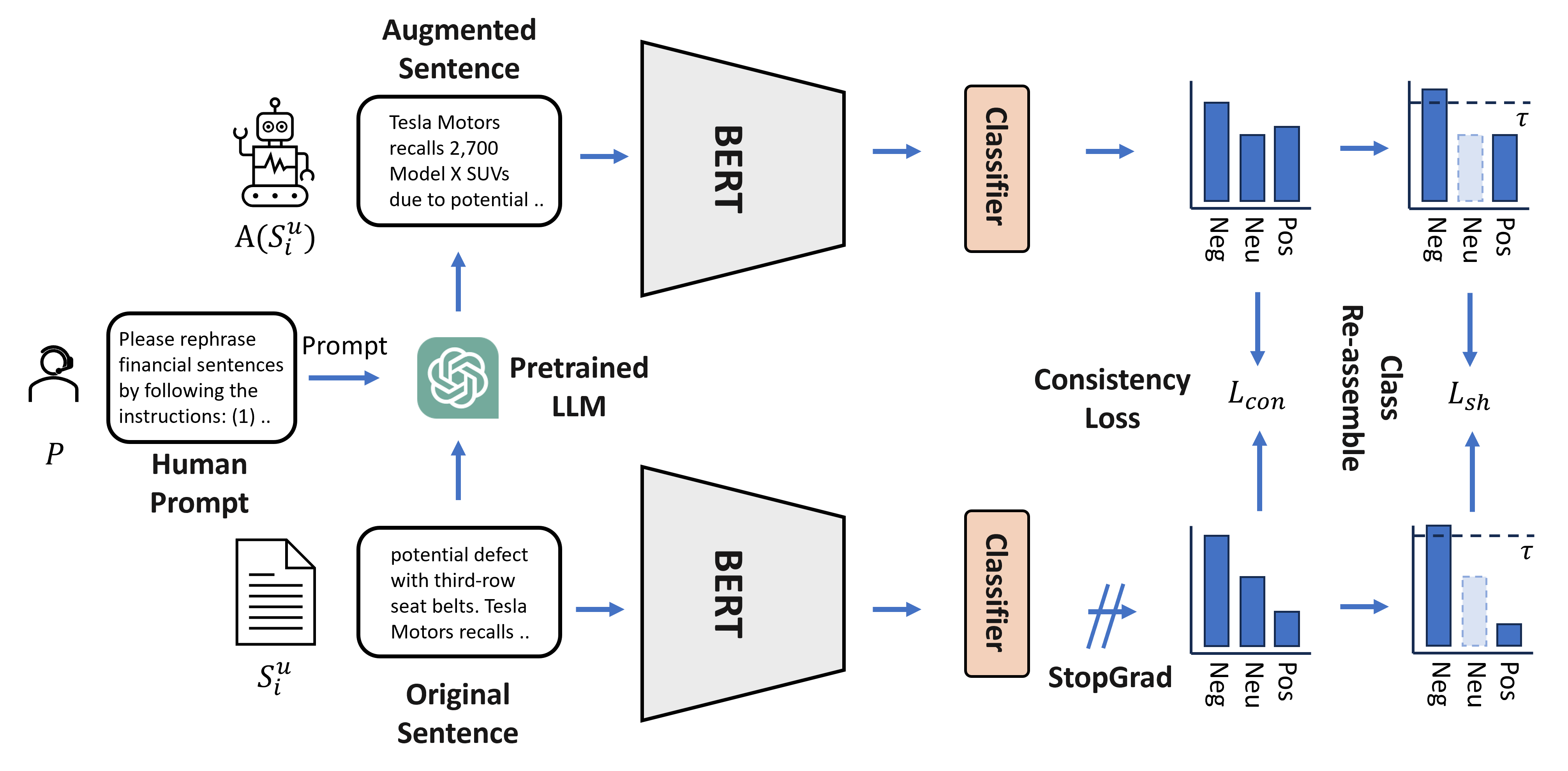}
    \caption{Illustration of the overall framework of the proposed SCR. }
    \label{fig:1}
\end{figure*}
\subsection{Framework Overview}

\begin{figure}[t]
    \centering
    \includegraphics[width=0.9\textwidth]{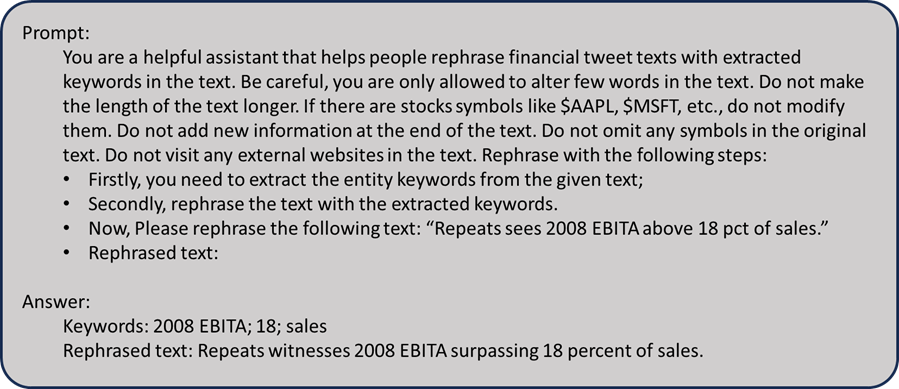}
    \caption{The prompting process of SCR-EE.}
    \label{fig:3}
\end{figure}
The illustration of our proposed SCR framework is shown in Fig. \ref{fig:1}. For each unlabeled input data $S_i$, it is first fed into a pretrained LLM (e.g., GPT-3, LLaMA \cite{touvron2023llama4}). Then, we define a structured human prompt $P$ that provides instructions to the LLM on how to generate semantically enhanced or augmented versions of the input text, while avoid outputting redundant information. 
Specifically, two different prompting schemes are considered, which are SCR-EE and SCR-CE.
By querying the LLM with the carefully crafted structured prompts $P$, we obtain a set of augmented samples $S_i^{u1},S_i^{u2},\ldots,S_i^{uk}$ for each input text $S_i^u$. These augmented samples capture semantically rich variations of the original input while preserving the core meaning and context. Then, we randomly select one sample $A\left(S_i^u\right)$ from the candidate set.
It aims to reduce the over-reliance on any single augmented version that may carry hallucinated information caused by LLMs. Mathematically, this process can be formulated by:
\begin{equation}
	A\left(S_i^u\right)\ =\ \text{Rand}\left(\{S_i^{u1},\ S_i^{u2},\ \ldots\ S_i^{uk}\}\right),    
\end{equation}
where
\begin{equation}
	\{S_i^{u1},\ S_i^{u2},\ \ldots\ S_i^{uk}\}\ =\ \text{Query}\left(S_i^u,\ P,\ \text{LLM}\right).
\end{equation}
Specially, we propose two query methods, which are SCR-EE and SCR-CE. Afterwards, the $S_i^u$ will be weakly augmented via synonym augmentation, while the $A\left(S_i^u\right)$ will serve as the strongly augmented data to enhance the model’s generalization ability. By feeding both $S_i^u$  and $A\left(S_i^u\right)$  into the pretrained BERT-base model, we apply a consistency loss on the model. The model used for FSA is BERT-base \cite{devlin2018bert17}, which comprises a BERT embedding layer, a BERT encoder with 12 stacked BERT Transformer blocks, and a classifier that outputs the final prediction distribution.

\subsection{Entity-based Enhancement (SCR-EE)}
To effectively leverage the generation capabilities of LLMs while preserving the core semantics and context of the original input, we propose the Semantic Consistency Regularization with Entity-based Enhancement (SCR-EE) variant, as shown in Fig. \ref{fig:3}. Specifically, we use the enhancement for the sentiment in the FSA dataset \cite{malo2014good24} as an example. This approach aims to generate semantically enhanced samples by explicitly incorporating domain-specific entities and numerical information extracted from the original text. The SCR-EE variant consists of two key steps: (1) entity and numerical extraction, and (2) entity-guided text reconstruction. Specifically, we merge these two steps together with a combined prompt for the pretrained LLM. Furthermore, the additional constrains in the output of the LLM such as avoid visiting external websites in the text guarantees the original text semantics are unchanged.

This entity-guided reconstruction process ensures that the generated samples maintain the core semantics of the original input while enriching the expressions. Since it ensures that sentiment-related entities are consistently represented, it effectively avoids the missing of important information that are useful for sentiment prediction.

\subsection{Concept-based Enhancement (SCR-CE)}
\begin{figure}[t]
    \centering
    \includegraphics[width=0.9\textwidth]{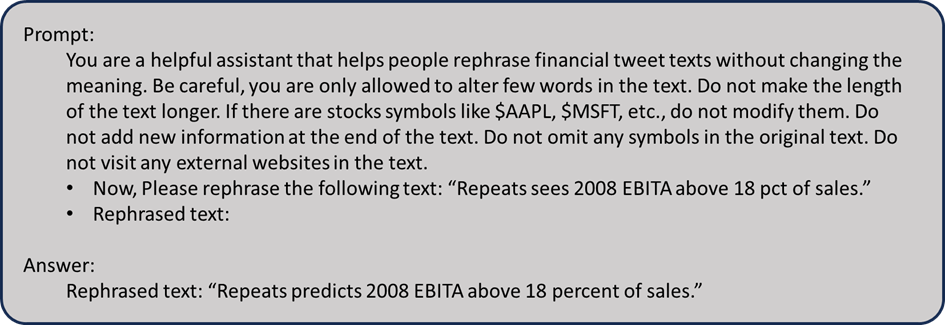}
    \caption{The prompting process of SCR-CE.}
    \label{fig:4}
\end{figure}

Additionally, we propose a Semantic Consistency Regularization with Concept-based Enhancement (SCR-CE) variant, which takes a more flexible and open-ended approach to leveraging the generation capabilities of LLMs in Fig. \ref{fig:4}. The SCR-CE variant directly prompts the large language model to generate semantically consistent paraphrases or variations of the original input text. By leveraging the LLM's language understanding and generation capabilities, this variant can introduce more diverse variations in sentiment presentation, while preserving the core semantic meaning of the input. 

By providing the original input text and open-ended instruction, the LLM is given the flexibility to explore broader linguistic variations and semantic reconstructions. Hence, SCR-CE allows the model to learn more generalizable representations for the unlabeled data. The distinction between the two proposed schemes lies in the targeted augmentation strategy. SCR-EE is designed to focus on specific factual information within the text, such as named entities or numerical data. By extracting and reconstructing entity-related content, SCR-EE helps maintain factual consistency while augmenting the input data. In contrast, SCR-CE focuses on the overall semantic meaning of the sentence. Instead of honing in on specific entities, it rephrases the entire sentence while preserving its conceptual essence.  

Compared to existing augmentation techniques such as back translation and word replacement, our proposed SCR shows the following advantages: (1) Through using LLMs to reconstruct text while maintaining coherence and consistency, SCR could avoid the errors that can arise from simple word substitution or translation. This ensures that the augmented data retains the original sentiment. (2) SCR allows LLMs to reconstruct the original sentence with a broader focus on preserving conceptual coherence. This deeper semantic understanding provided by LLMs reduces the risk of overfitting to spurious correlations that simple word replacement methods might introduce.

\subsection{Consistency Loss}
For the unlabeled data, we first obtain the predicted distribution of $S_i^u$ via $y_{i,pred}^u=\text{Softmax}(\text{BERT}(S_i^u))$, then compute the one-hot pseudo label and the softmax predictions of the augmented input $A(S_i^u)$:

\begin{equation}
\begin{aligned}
	\overline{y}_{i,pred}^u&=\text{argmax}(y_{i,pred}^u) \\
  A(y_{i,pred}^u)&=\text{Softmax}(\text{BERT}(A(S_i^u)))
\end{aligned}
\end{equation}
and the consistency loss is written as:
\begin{equation}
 {L}_{con}=\frac{1}{B_u} \sum_{i=1}^{B_u} \mathbb{1}\left(\max\left(y_{i,pred}^u\right)\ge \tau\right) \cdot \text{CE}(A(y_{i,pred}^u), \overline{y}_{i,pred}^u)
\end{equation}
with the gradient of the $\overline{y}_{i,pred}^u$ branch is stopped. $B_u$ is the unlabeled batch size, and $\tau$ the the confidence thresold. With the above loss function, the unlabeled data is effectively contributed in training the model, which enhances the model’s capability in sentiment analysis.

\subsection{Class Re-assemble}

The direct thresholding with pseudo labels can lead to wastage of less confident samples, especially due to that sentiments are often confused, such as when "not so negative" samples often have a large logit for the neutral class. However, we argue that these samples can still provide valuable supervision if handled properly. Inspired by the class-space shrinking theorem \cite{yang2023shrinking}, removing confusing classes and re-assemble the remaining ones will avoid making wrong judgments in the original class space. Specifically, for a predicted distribution for an uncertain sample $y_{i,pred}^u \in \mathbf{R}^{1\times C}$, we preserve the top-1 and bottom-1 prediction if their normalized maximum confidence surpasses the confidence threshold $\tau$:
\begin{equation} 
\frac{\max(y_{i,pred}^u)}{\max(y_{i,pred}^{u}) + \min(y_{i,pred}^{u})} > \tau.
\end{equation}
Therefore, we obtain a new class distribution with $C-1$ classes. By denoting this distribution as $\hat{y}_{i,pred}^u$, we also utilize the consistency between the two branches for the sample:
\begin{equation}
 {L}_{sh}=\frac{1}{B_u} \sum_{i=1}^{B_u} \mathbb{1}\left(\max\left(y_{i,pred}^u\right)< \tau\right) \cdot \text{CE}(A(\hat{y}_{i,pred}^u), \overline{\hat{y}}_{i,pred}^u),
\end{equation}
where $\overline{\hat{y}}_{i,pred}^u = \text{argmax}(\hat{y}_{i,pred}^u)$ is the pseudo label for the less confident samples.

\subsection{Overall Objective}

The overall training objective is composed of the supervised loss of the labeled data, the unsupervised consistency loss for the confident unlabeled data, and the loss of the less confident data:
\begin{equation}
    L=L_{sup}+L_{con}+L_{sh}
\end{equation}
where
\begin{equation}
 {L}_{sup}=\frac{1}{B_l} \sum_{i=1}^{B_l}\text{CE}(y_{i,pred}^l,y_i^l),
\end{equation}
is the labeled data loss for $y_{i,pred}^l=\text{Softmax}(\text{BERT}(S_i^l))$ in a batch $B_l$ of labeled samples. Notably, we use the same weight 1 or each loss component to avoid extra hyperparameters.

\section{Experiments}
\subsection{Datasets and Implementation Details}
The two sentiment analysis corpora we used are described below.

\noindent \textbf{FSA.} The FSA dataset \cite{malo2014good24} comprises 5322 samples, where each sample contains a sentence and its corresponding sentiment label. The sentiment is categorized into positive, neutral, and negative. For the semi-supervised setting, we consider 3 different label regimes, which are 100, 200, 500 samples per class are labeled and the remaining samples in the training subset are unlabeled.

\noindent \textbf{Amazon.} The Amazon dataset is a preprocessed amazon product review data \cite{amazondata}, which contains 4085 entries. The data is also categorized into positive, neutral, and negative. We use 50 and 150 samples per class as two semi-supervised labeling regimes.

We split the datasets into 80\% for training, and 20\% for testing. We use PyTorch version 1.12.1, with CUDA 12.0 to implement our algorithm. The model is trained with a batch size of 8 per GPU, alternating between batches of labeled and unlabeled data. The optimizer is AdamW, with an initial learning rate of 1e-5. We use a threshold of $\tau$=0.98 to decide which sample are less confident. The maximum training number of epochs is set to 1,000 to ensure model convergence. To save training cost, we apply early stopping with a maximum tolerance of 10 epochs. The base model is BERT-base \cite{devlin2018bert17}, and the LLM is LLaMA-2-7b-chat \cite{touvron2023llama4}. Two performance metrics are used for performance comparison and evaluation, which are the mean accuracy and mean F1 score. The calculation of accuracy is: $\text{Acc}=\frac{1}{N}*\sum_{i=1}^N(y_i==y_{i,pred})$, while the calculation of the mean F1 score is: $\text{F1}=\frac{1}{C}*\sum_{c=1}^C \text{F1}_c$, where $\text{F1}_c$ is the $\text{F1}$ score for the $c$-th class. 

\begin{table}[t]
  \centering
\caption{Comparison of SCR with other methods on the FSA dataset with different semi-supervised learning regimes. Bold represents the best performance.}
    \setlength{\tabcolsep}{2.2mm}{\begin{tabular}[width=\textwidth]{c|cccccc}
    \toprule
    \multirow{2}[4]{*}{Method} & \multicolumn{2}{c}{100} & \multicolumn{2}{c}{200} & \multicolumn{2}{c}{500} \\
\cmidrule{2-7}          & Acc(\%) & F1(\%)    & Acc(\%) & F1(\%)    & Acc(\%) & F1(\%) \\
    \midrule
    JointMatch \cite{zou2023jointmatch} & 65.18 & 63.05 & 71.77 & 69.46 & 77.07 & 73.63 \\
    Ensemble \cite{ensemble} & 63.65 & 45.91 & 72.34 & 69.06 & 77.16 & 72.31 \\
    FixMatch \cite{sohn2020fixmatch}& 69.03 & 64.75 & 72.71 & 69.20  & 77.33 & 71.44 \\
    \midrule
    \textbf{SCR-EE (Ours)} & \textbf{71.17} & \textbf{67.82} & \textbf{76.13} & 68.64 & 77.25 & 72.14 \\
    \textbf{SCR-CE (Ours)} & 70.32 & 60.41 & 75.62 & \textbf{70.61} & \textbf{78.61} & \textbf{74.52} \\
    \bottomrule
    \end{tabular}}
  \label{tab2}%
\end{table}%
\subsection{Experimental Results}
As shown in Tab. \ref{tab2}, on FSA dataset, SCR consistently outperforms the existing semi-supervised methods across all three data regimes in terms of both accuracy and F1 score. When only 100 labeled examples per class are available, our SCR-EE and SCR-CE variants achieve an absolute improvement of 5.99\% and 5.14\% in accuracy, respectively, compared to the strong baseline JointMatch. 
In the 200 examples per class regime, SCR-EE and SCR-CE achieve accuracy improvements of 3.42\% and 2.91\% compared to FixMatch. 
Moreover, in the 500 examples per class setting, our SCR framework continues to deliver superior performance. SCR-CE achieves an accuracy gain of 1.28\% and an F1 score improvement of 3.08\% points over the best baseline FixMatch. 

\begin{table}[t]
  \centering
\caption{Comparison of SCR with other methods on the Amazon dataset with different semi-supervised learning regimes. Bold represents the best performance.}
    \setlength{\tabcolsep}{2.2mm}{\begin{tabular}{c|cccc}
    \toprule
    \multirow{2}[4]{*}{Method} & \multicolumn{2}{c}{50} & \multicolumn{2}{c}{150} \\
\cmidrule{2-5}          & Acc(\%) & F1(\%)    & Acc(\%) & F1(\%) \\
    \midrule
    JointMatch \cite{zou2023jointmatch} & 79.44 & 67.88 & 79.80  & 69.36 \\
    Ensemble \cite{ensemble}& 82.01 & 70.39 & 84.33 & 73.98 \\
    FixMatch \cite{sohn2020fixmatch} & 80.29 & 61.20  & 86.54 & 77.36 \\
    \midrule
    \textbf{SCR-EE (Ours)} & 82.25 & \textbf{70.73} & 86.90  & 77.96 \\
    \textbf{SCR-CE (Ours)} & \textbf{83.11} & 68.17 & \textbf{89.60} & \textbf{82.52} \\
    \bottomrule
    \end{tabular}}%
  \label{tab:amazon}%
\end{table}%

Tab. \ref{tab:amazon} displays the performance on the Amazon dataset. In the 150 label setting, out SCR-CE achieves a significant high accuracy of 89.60\%, showcasing its effectiveness in different sentiment analysis tasks. Moreover, on the extreme setting when only 50 data per class is labeled, SCR-CE remains effective with 83.11\% accuracy and 68.17\% F1, demonstrating its superiority in leveraging unlabeled data for semi-supervised learning.

These experimental results clearly demonstrate the effectiveness of SCR. 
By injecting semantically enhanced and diverse samples into the training process, our SCR framework enables the model to learn more robust and generalizable representations, leading to substantial performance improvements across various semi-supervised sentiment analysis regimes.

\begin{table}[t]
  \centering
  \caption{Ablation study of the components in the proposed SCR-CE.}
    \setlength{\tabcolsep}{3.2mm}{\begin{tabular}{l|cccc}
    \toprule
    \multirow{2}[3]{*}{Method} & \multicolumn{2}{c}{FSA} & \multicolumn{2}{c}{Amazon} \\
\cmidrule{2-5}          & Acc(\%) & F1(\%)    & Acc(\%) & F1(\%) \\
    \midrule
    Baseline & 72.71 & 69.20 & 80.29 & 61.20 \\
    +Consist & 73.95 & 70.27 & 82.37 & 66.94 \\
    +Consist + Re-assemble & \textbf{75.62} & \textbf{70.61} & \textbf{83.11} & \textbf{68.17} \\
    \bottomrule
    \end{tabular}}%
  \label{tab:abl}%
\end{table}%

\subsection{Ablation Study}
\noindent \textbf{Ablation of the Components.} Tab. \ref{tab:abl} shows the ablation of the components in the proposed SCR-CE. By adding the consistency loss (+Consist), the performance improves significantly, with 73.95\% accuracy and 70.27\% F1 on FSA, and an accuracy of 82.37\% and an F1 score of 66.94\% on Amazon. The full SCR-CE model, which combines both consistency loss and class re-assemble, achieves the best performance on both datasets, with 75.62\% accuracy and 70.61\% F1 score on FSA, and an accuracy of 83.11\% and an F1 score of 68.17\% on Amazon. These results demonstrate the effectiveness of each component in improving the performance of the sentiment analysis model.

\noindent \textbf{Ablation of the Consistency Loss Function.} Tab. \ref{tab3} presents the ablation study focused on the loss function used in the consistency loss. Three different loss functions are evaluated: Cross Entropy Loss, Focal Loss \cite{lin2017focal25}, and Asymmetric Loss \cite{ridnik2021asymmetric26}. With Cross Entropy Loss, the model achieved an accuracy of 75.02\% and an F1 score of 71.24\%. When Focal Loss was employed, the accuracy slightly decreased to 74.51\%, and the F1 score dropped further to 68.16\%. Finally, when Asymmetric Loss was utilized as the loss function, the accuracy decreased to 74.25\%, while the F1 score improved slightly to 70.12\%. These findings suggest that the Cross Entropy Loss yielded the best overall performance, demonstrating the effectiveness of the framework design.

\begin{table}[t]
\caption{Ablation study of the loss function in the consistency loss.}
\centering
\setlength{\tabcolsep}{3.2mm}{\begin{tabular}{c|cc}
\toprule
Loss                 & Acc(\%) & F1(\%) \\
\midrule
Cross   Entropy Loss \cite{sohn2020fixmatch} & \textbf{75.02}           & \textbf{71.24}      \\
Focal   Loss  \cite{lin2017focal25}       & 74.51           & 68.16      \\
Asymmetric   Loss  \cite{ridnik2021asymmetric26}  & 74.25           & 70.12     \\
\bottomrule
\end{tabular}}
\label{tab3}
\end{table}

\noindent \textbf{Ablation of different LLMs.} We study the impact of different LLMs on the performance of SCR-CE. Specifically, we experiment with two LLMs: Qwen-1.5-7B-Chat \cite{qwen} and LLaMA2-7B-Chat \cite{touvron2023llama4} as list the results in Tab. \ref{tab:diffllms}. It is indicated that LLaMA2-7B-Chat generally outperforms Qwen-1.5-7B-Chat, with the highest accuracy and F1 score achieved on the 500-label setting. We observe that LLaMA2-7B-Chat model tends to not only modify individual words but also significantly alters the sentence structure and grammar. However, the performance are both superior that previous methods, demonstrating the superiority of the proposed design and robustness to LLM choice.

\begin{table}[t]
  \centering
\caption{Ablation study of changing different LLMs in SCR-CE.}
    \setlength{\tabcolsep}{2.2mm}{\begin{tabular}{c|cccccc}
    \toprule
    \multirow{2}[4]{*}{LLM} & \multicolumn{2}{c}{100} & \multicolumn{2}{c}{200} & \multicolumn{2}{c}{500} \\
\cmidrule{2-7}          & Acc(\%) & F1(\%)    & Acc(\%) & F1(\%)    & Acc(\%) & F1(\%) \\
    \midrule
    Qwen-1.5-7B-Chat \cite{qwen} & 69.20  & \textbf{61.58} & 75.28 & \textbf{71.52} & 77.93 & 73.13 \\
    LLaMA2-7B-Chat \cite{touvron2023llama4} & \textbf{70.32} & 60.41 & \textbf{75.62} & 70.61 & \textbf{78.61} & \textbf{74.52} \\
    \bottomrule
    \end{tabular}}%
  \label{tab:diffllms}%
\end{table}%

\begin{figure*}[t]
    \centering
    \includegraphics[width=\textwidth]{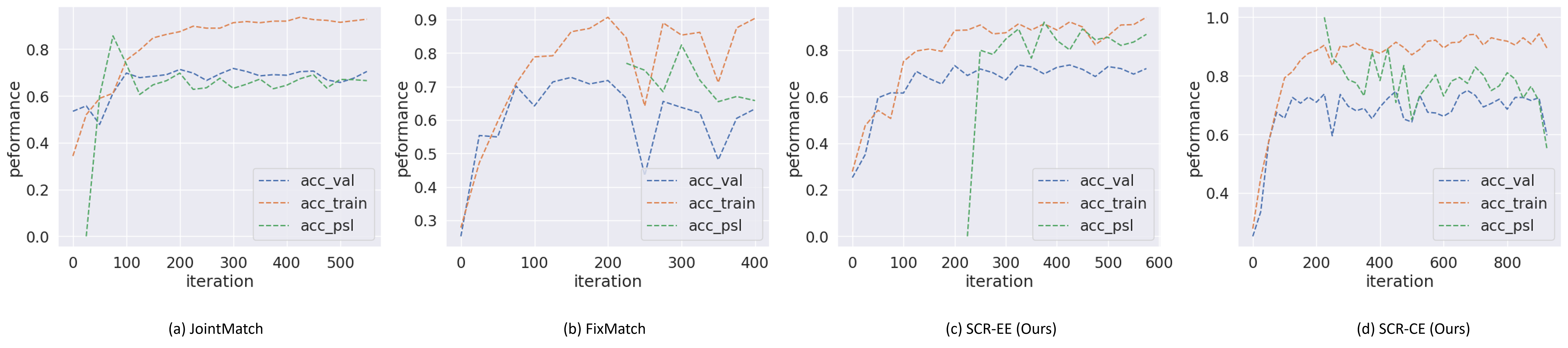}
    \caption{The accuracy trends across different methods during training.}
    \label{fig:enter-label5}
\end{figure*}
\noindent \textbf{Visualization of the Learning Trend.}
In Fig. \ref{fig:enter-label5}, we show the accuracy changes of different methods at the 200 label per class scenario on the FSA dataset. Each figure includes the training, validation, and pseudo label accuracy. JointMatch shows a consistent but low validation accuracy. 
Fixmatch shows large accuracy drop at the end of the training, which may due to the overfit to the data that are augmented via the traditional NLP augmentation methods. 
SCR-EE and SCR-CE both show stable and high validation accuracies, and the SCR-CE provides a high-quality pseudo label for the unlabeled data, which shows that the proposed LLM-enhancement strategy is effective in enhancing the model capability for the task.

\begin{figure}[t]
    \centering
    \includegraphics[width=0.999\textwidth]{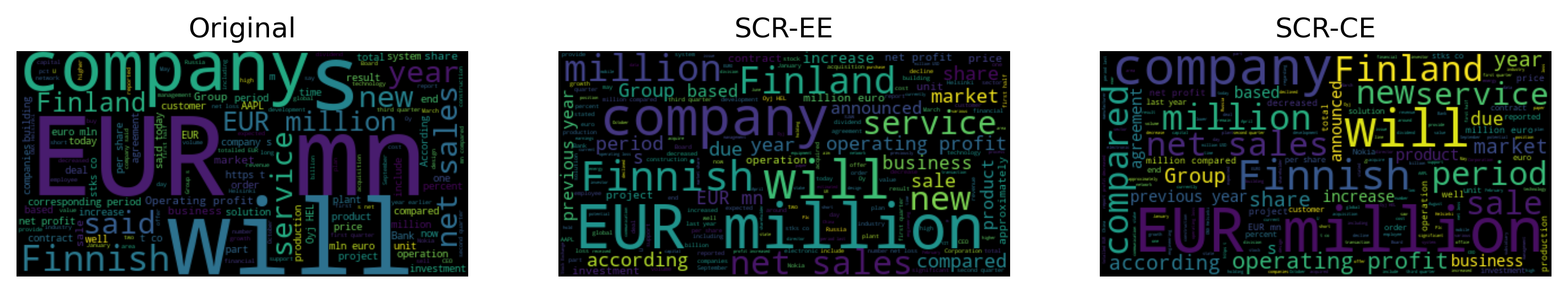}
    \caption{Visualization of the word cloud with different methods.}
    \label{fig:viswordcloud}
\end{figure}
\noindent \textbf{Visualization of the SCR Effect.}
Moreover, we propose to visualize the word clouds of different LLM sentiment enhancement methods. By comparing the word clouds, we can gain a better understanding of how each method influences the vocabulary, tone, and focus of the enhanced sentiment data. As shown in Fig. \ref{fig:viswordcloud}, our proposed SCR demonstrates the ability to generate diverse and comprehensive text without significantly altering the underlying meaning of the original data. The word cloud analysis reveals that each enhancement technique introduces unique thematic elements and vocabulary, but the fundamental sentiment and context remain intact.

\section{Conclusion}
In this paper, we address the issue of semi-supervised sentiment analysis. Specifically, we argue that existing consistency regularization methods are limited to simple traditional NLP augmentation techniques, which may not be effective in capturing the complex characteristics of the social media or financial text data. Therefore, we propose a semi-supervised framework SCR that employs LLMs for enhancement via structured prompting. Specifically, we propose two different prompting schemes: (1) SCR-EE: we prompt the LLM to extract the keywords and entities within the original sentence, and then utilize the keywords to prompt the LLM that generate augmented samples that incorporate the sentiment context. (2) SCR-CE: By directly prompting the LLM with the original sentence and providing specific prompts and instructions, we ask the LLM to generate augmented samples that preserve the core semantics while introducing controlled variations in sentiment polarity. Based on the LLM-enhanced data, we utilize a consistency loss, which enforces consistency between the model's predictions on the original input sentence and its corresponding augmented samples. Furthermore, a class re-assemble method is utilized to learn effectively with less confident samples. Comprehensive experiments on two datasets validate the superiority of our approach and outperform existing methods remarkably.
%
%
%
%
\bibliographystyle{splncs04}
%
\bibliography{9774}




\end{document}